\documentclass[11pt,a4paper]{article}
\usepackage[hyperref]{emnlp-ijcnlp-2019}
\usepackage{times}
\usepackage{latexsym}
\usepackage{booktabs}
\usepackage{url}
\usepackage{linguex}
\usepackage{graphicx}
\usepackage{subcaption}
\usepackage{xcolor,colortbl}
\usepackage{multirow}
\usepackage{mathtools}
\usepackage{amsfonts}

\aclfinalcopy 

\definecolor{light-gray}{gray}{0.8}
\definecolor{dark-gray}{gray}{0.5}

\title{Using Priming to Uncover the Organization of Syntactic Representations in Neural Language Models}

\author{Grusha Prasad \\
  Johns Hopkins University  \\
  {\tt grusha.prasad@jhu.edu}\\
   \\
   \\
   \And
  Marten van Schijndel \\
  Cornell University  \\
  {\tt mv443@cornell.edu} \\
 \\
  \\
   \And
  Tal Linzen \\
  Johns Hopkins University  \\
  {\tt tal.linzen@jhu.edu} \\
   \\
   \\
  }

\date{}

\begin{document}
\maketitle
\begin{abstract}
    Neural language models (LMs) perform well on tasks that require sensitivity to syntactic structure. Drawing on the syntactic priming paradigm from psycholinguistics, we propose a novel technique to analyze the representations that enable such success. By establishing a gradient similarity metric between structures, this technique allows us to reconstruct the organization of the LMs' syntactic representational space. We use this technique to demonstrate that LSTM LMs' representations of different types of sentences with relative clauses are organized hierarchically in a linguistically interpretable manner, suggesting that the LMs track abstract properties of the sentence.  
\end{abstract}

\setlength{\Exlabelwidth}{0.75em}
\setlength{\SubExleftmargin}{1.35em}

\section{Introduction}
Neural networks trained on text alone, without explicit syntactic supervision, have been surprisingly successful in tasks that require sensitivity to sentence structure. The difficulty of interpreting the learned neural representations that underlie this success has motivated a range of analysis techniques, including diagnostic classifiers \cite{giulianelli18, conneau18, shi16}, visualization of individual neuron activations \cite{kadar17, qian16}, ablation of individual neurons or sets of neurons \cite{lakretz19} and behavioral tests of generalization to infrequent or held out syntactic structures \cite{linzen16,weber18,mccoy18}; for reviews, see \citet{belinkov19} and  \citet{alishahi19}. 

This paper expands the toolkit of neural network analysis techniques by drawing on the \textbf{syntactic priming} paradigm, a central tool in psycholinguistics for analyzing human syntactic representations \cite{bock1986syntactic}. This paradigm is based on the empirical finding that people tend to reuse syntactic structures that they have recently produced or encountered. For example, English provides two roughly equivalent ways to express a transfer event:

\ex.\label{ex:dative}
\a.The boy threw the ball to the dog.\label{ex:dativepo1}
\b.The boy threw the dog the ball.\label{ex:dativedo1}
  
When readers encounter one of these variants in the text more frequently than the other, they expect that future transfer events will more likely be expressed using the frequent construction than the infrequent one. For example, after reading sentences like \ref{ex:dativepo1} (the \textbf{prime}), readers expect sentences like \ref{ex:dativepo2}, which shares syntactic structure with the prime, to occur with a greater likelihood than the alternative variant like \ref{ex:dativedo2} which does not \cite{wells09}.\footnote{\Citet{wells09} measured priming effects for relative clauses, not dative constructions. For work on priming in production with dative constructions, see \citet{kaschak11}.}


\ex.
\a. The lawyer sent the letter to the client. \label{ex:dativepo2}
\b. The lawyer sent the client the letter. \label{ex:dativedo2}


We use the priming paradigm to analyze neural network language models (LMs), systems that define a probability distribution over the $n$\textsuperscript{th} word of a sentence given its first $n-1$ words. Building on paradigms that determine whether the LM's expectations are consistent with the syntactic structure of the sentence \cite{linzen16}, we measure the extent to which a LM's expectation for a specific syntactic structure is affected by recent experience with related structures. We prime a fully trained model with a structure by adapting it to a small number of sentences containing that structure \citep{vanSchijndel18adaptation}. We then measure the change in surprisal (negative log probability) after adaptation when the LM is tested either on sentences with the same structure or sentences with different but related structures. The degree to which one structure primes another provides a graded similarity metric between the model's representations of those structures (cf. \citealt{branigan2017experimental}), which allows us to investigate how the representations of sentences with these structures are organized.


As a case study, we applied this technique to investigate how recurrent neural network (RNN) LMs represent sentences with relative clauses (RCs). We found that the representations of these sentences are organized in a linguistically interpretable manner: sentences with a particular type of RC were most similar to other sentences with the same type of RC in the LMs' representation space. Furthermore, sentences with different types of RCs were more similar to each other than sentences without RCs. We demonstrate that the similarity between sentences was not driven merely by specific words that appeared in the sentence, suggesting that the LMs tracked abstract properties of the sentence. This ability to track abstract properties decreased as the training corpus size increased. Finally, we tested the hypothesis that LMs' accuracy on agreement prediction \citep{marvin18} would increase with the LMs' ability to track more abstract properties of the sentence, but did not find evidence for this hypothesis. 

\begin{table*}
    \resizebox{\textwidth}{!}{
		\begin{tabular}{ll}
		\toprule
		\textbf{Abstract structure} & \textbf{Example} \\
		\midrule
		Unreduced Object RC & The conspiracy that the employee welcomed divided the \textcolor{gray}{beautiful} country. \\
		Reduced Object RC & The conspiracy the employee welcomed divided the \textcolor{gray}{beautiful} country. \\
		Unreduced Passive RC & The conspiracy that was welcomed by the employee divided the \textcolor{gray}{beautiful} country. \\
		Reduced Passive RC & The conspiracy welcomed by the employee divided the \textcolor{gray}{beautiful} country.\\
		Active Subject RC & The employee that welcomed the conspiracy \textcolor{gray}{quickly} searched the building\textcolor{gray}{s}. \\
		PS/ORC-matched Coordination & The conspiracy welcomed the employee and divided the \textcolor{gray}{beautiful} country. \\
		ASRC-matched Coordination & The employee welcomed the conspiracy and \textcolor{gray}{quickly} searched the building\textcolor{gray}{s}.\\
		\bottomrule

		\end{tabular}
		}
		\caption{\label{tab:ex} Examples of sentences generated using templates containing the seven abstract structures we analyzed (optional elements, which only occur in a subset of the examples, are indicated in grey).}
\end{table*}

\section{Background}

\subsection{Syntactic predictions in neural LMs}

We build on paradigms that use LM probability estimates for words in a given context as a measure of the model's sensitivity to the syntactic structure of the sentence \cite{linzen16, gulordava18, marvin18}. If a language model assigns a higher probability to a verb form that agrees in number with the subject (\textit{the boy... writes}) than a verb form that does not (\textit{the boy... write}), we can infer that the model encodes information about the agreement features of nouns and verbs (that is, the difference between singular and plural) and has correctly identified the subject that corresponds to this verb. This reasoning has been extended beyond subject-verb agreement to study whether the predictions of neural LMs are sensitive to a range of other syntactic dependencies, including negative polarity items \cite{jumelet18}, filler-gap dependencies \cite{wilcox18} and reflexive pronoun binding \cite{futrell19}.

\subsection{Syntactic priming in humans \label{sec:human_priming}} 
Syntactic priming has been used to study whether the representations of two sentences have shared structure. For example, \ref{ex:dativepo1} (repeated below as \ref{ex:dativepo_repeat}) shares the structure VP~$\rightarrow$ V NP PP with \ref{ex:dativepo3} but not \ref{ex:dativedo3}. 

\ex. \label{ex:dativepo_repeat} The boy threw the ball to the dog.

\ex. 
\a. \label{ex:dativepo3} The renowned chef made some wonderful pasta for the guest.
\b. \label{ex:dativedo3} The renowned chef made the guest some wonderful pasta. 

If \ref{ex:dativepo_repeat} primes \ref{ex:dativepo3} more than it primes \ref{ex:dativedo3}, we can infer that the representations of \ref{ex:dativepo_repeat} are more similar to that of \ref{ex:dativepo3} than to that of \ref{ex:dativedo3}. Since \ref{ex:dativedo3} and \ref{ex:dativepo3} differ only in their structure, this difference in similarity must be driven by structural information in the representations of the sentences (for reviews, see \citealt{mahowald16} and \citealt{tooley10}).

Although priming studies have traditionally measured the priming effect on the sentence immediately following the prime, more recent studies have demonstrated that the effects of syntactic priming can be cumulative and long-lasting: sentences with a shared structure $S_X$ become progressively easier to process when preceded by $n$ sentences with the same structure $S_X$ than when preceded by $n$ sentences with a different structure $S_Y$ \cite{kaschak11, wells09}.\footnote{In studies looking at non-cumulative priming, $n=1$.} In conjunction with the finding that words that are consistent with a probable syntactic parse are easier to process than words consistent with less probable parses \cite{hale01, levy08}, the increased ease of processing in cumulative priming studies can be interpreted as evidence that, with increased exposure to a structure, participants begin to expect that structure with a greater probability \cite{chang06}.

Cumulative priming allows us to study how sentences are related to each other in the human (or LM) representation space in the same way that non-cumulative priming does: when participants (or LMs) are exposed to sentences with structure $S_X$, if there is a greater decrease in surprisal when they are tested on other sentences with $S_X$ than when they are tested on other sentences with $S_Y$, we can infer that the representations of sentences with $S_X$ are more similar to each other than to the representations of sentences with $S_Y$.

\subsection{LM adaptation as cumulative priming \label{sec:adaptation_as_priming}}
\Citet{vanSchijndel18adaptation} modeled cumulative priming in recurrent neural networks (RNNs) by adapting fully trained RNN LMs to new stimuli  --- i.e. taking a fully trained RNN LM and continuing to train it on a small set of sentences (cf.\ \citealt{graveetal17,krauseetal17, chowdhury19}).  They demonstrated that when an RNN LM was adapted to a small number of sentences with a shared syntactic structure, the surprisal for novel sentences with that structure decreased, enabling them to infer that the LM's representations of sentences contained information about that structure.  

\section{Similarity between syntactic structures in RNN LM representational space} \label{sec:adapt(y|x)}

Following the assumptions in Section~\ref{sec:human_priming}, we define a similarity metric between two structures $S_X$ and $S_Y$ in an LM's representation space by adapting the LM to sentences with $S_X$ and measuring the change in surprisal for sentences with $S_Y$ --- i.e. measuring to what extent sentences with $S_X$ prime sentences with $S_Y$. We use the notation $\mathbb{A}(Y \mid X)$ to refer to this change in surprisal\footnote{ $\mathbb{A}$ is shorthand for adaptation.}, where $X$ and $Y$ are non-lexically-overlapping sets of sentences whose members share the structures $S_X$ and $S_Y$ respectively. If we assume that $S_X$ and $S_Y$ are similar to each other in the LM's representation space, then $\mathbb{A}(Y \mid X) > 0$ --- i.e., encountering sentences with $S_X$ causes the LM to assign a higher probability to sentences with $S_Y$. On the other hand, if we assume that $S_X$ and $S_Y$ are unrelated to each other, then $\mathbb{A}(Y \mid X) = 0$ --- i.e., encountering sentences with $S_X$ does not cause the LM to change its probability for sentences with $S_Y$. 

\section{Experimental setup}

\subsection{Syntactic structures}
We analyzed five types of RCs. In an active subject RC, the gap is in the subject position of the embedded clause:\footnote{We illustrate the location of the gap with underscores here, but the underscores were not included in the LM's input.}

\ex. \label{ex:src} My cousin that $\rule{0.17cm}{0.15mm}$ liked the book ...

In a passive subject RC (\textit{passive RCs}), the gap is in the subject position of the embedded clause, and the embedded verb is passive. In English, passive RCs can be unreduced \ref{ex:unreduced_prc} or reduced \ref{ex:reduced_prc}:

\ex.\label{ex:prc}\a.The book that $\rule{0.17cm}{0.15mm}$ was liked by my cousin ...\label{ex:unreduced_prc}
\b.The book $\rule{0.17cm}{0.15mm}$ liked by my cousin ...\label{ex:reduced_prc}

In an object RC the gap is in the object position of the embedded clause. In English, object RCs can be unreduced \ref{ex:unreduced_orc} or reduced \ref{ex:reduced_orc}:

\ex. \label{ex:orc} \a. The book that my cousin liked $\rule{0.17cm}{0.15mm}$ ... \label{ex:unreduced_orc}
\b.The book my cousin liked $\rule{0.17cm}{0.15mm}$ ... \label{ex:reduced_orc}

Finally, we also included two additional conditions with verb coordination: one with nearly identical word order and lexical content as active subject RCs (\ref{ex:scoord}; ASRC-matched Coordination), and another with nearly identical word order and lexical content as passive RCs and object RCs (\ref{ex:ocoord}; PS/ORC-matched Coordination).\footnote{In order to maintain the same word order as in object and passive RCs, the subject of the coordinated verb phrases is an NP that tends to fill the object position in other sentences (e.g, ``the equation"). Therefore, many of the sentences in this condition are implausible (e.g., ``The equation reviewed the physicists and challenged the method.") \label{footnote:coord_semantics}}

\ex. My cousin liked the book and ... \label{ex:scoord}

\ex. The book liked my cousin and ... \label{ex:ocoord}

These conditions enable us to measure whether sentences with different types of RCs are more similar to each other in an LM's representation space than they are to lexically matched sentences without RCs. 

\begin{figure}
    \centering
    \includegraphics[width = 0.43\textwidth]{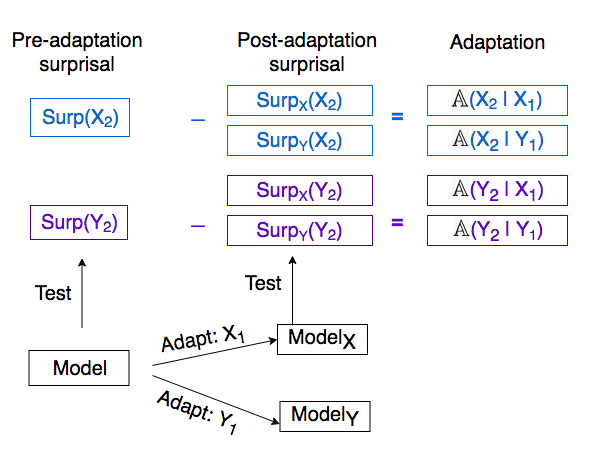}
    \caption{A schematic for calculating the similarity between two structures $S_X$ and $S_Y$ in an LM's representation space. $X_1$, $X_2$ and $Y_1$, $Y_2$ are non-lexically-overlapping sets of sentences with $S_X$ and $S_Y$ respectively. $\textit{Model}_X$ and $\textit{Model}_Y$ refer to versions of a fully trained model that have been adapted to either $X_1$ or $Y_1$ respectively. $\textit{Surp}_X()$ and $\textit{Surp}_Y()$ are functions that return the surprisal of sentences for $\textit{Model}_X$ and $\textit{Model}_Y$.}
    \label{fig:exp_schema}
\end{figure}

\subsection{Adaptation and test sets}
We generated sentences from seven templates, one for each of the syntactic structures of interest. The slots were filled with 223 verbs, 164 nouns, 24 adverbs and 78 adjectives such that the semantic plausibility of the combination of nouns, verbs, adverbs and adjectives was ensured. The seven variants of every sentence had nearly identical lexical items (see Table~\ref{tab:ex}).%
\footnote{Since the main verb of the sentence was constrained to be semantically plausible with the subject of the sentence, it often varied between active subject RC and ASRC-matched coordination on the one had and all other conditions on the other.}
We used these templates to generate five experimental lists --- each list comprised of a pair of adaptation and test sets with minimal lexical overlap between them (only function words and some modifiers were shared). Each adaptation set contained 20 sentences and each test set contained 50.%

In order to infer that any decrease in surprisal is caused by adaptation to an abstract syntactic structure, we need to ensure that the models are not adapting to properties of the sentence that are unrelated to the abstract structure of interest. Consider a LM adapted to \ref{ex1} and tested on \ref{ex2}:

\ex. \label{ex1} The conspiracy that the employee welcomed divided the country.

\ex. \label{ex2} The proposal that the receptionist managed shocked the CEO. 

\begin{figure*}
    \centering
    \begin{subfigure}[b]{0.5\textwidth}
    \centering
        \includegraphics[width=\textwidth]{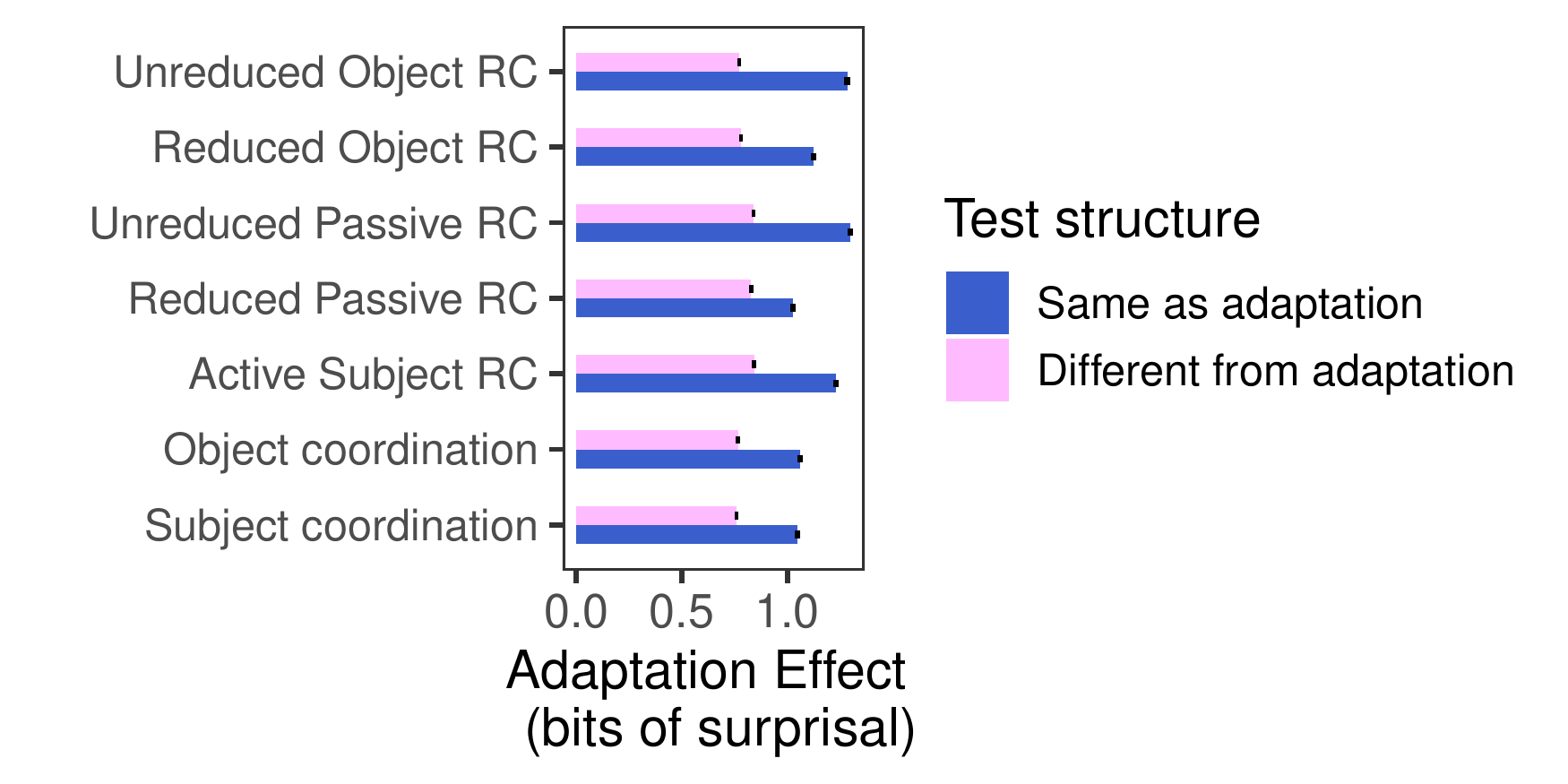}
        \caption{}
        \label{fig:analysis1}
    \end{subfigure}\hfill
    \begin{subfigure}[b]{0.5\textwidth}
        \includegraphics[width=0.95\textwidth]{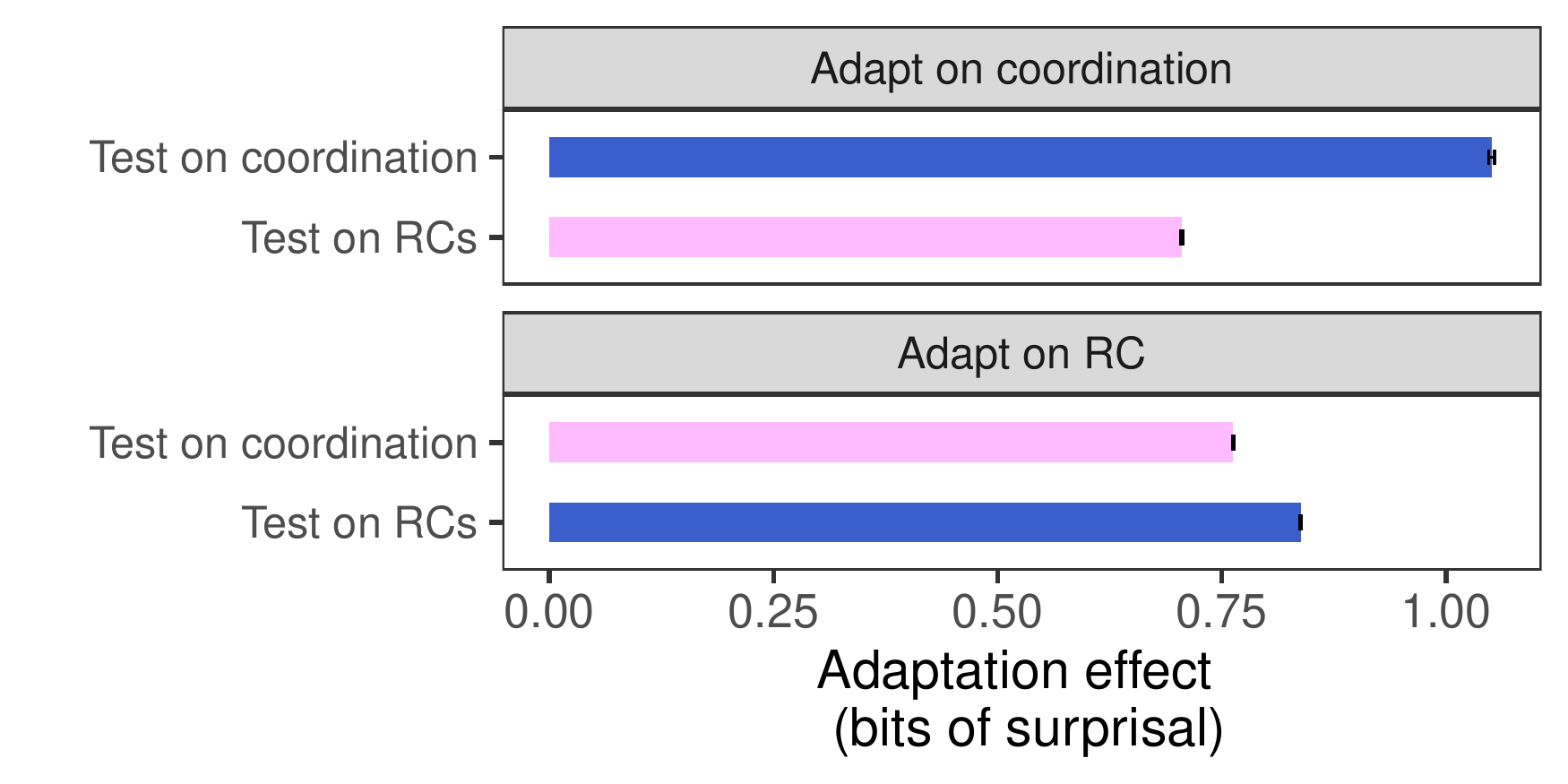}
        \caption{}
        \label{fig:analysis2}
    \end{subfigure}
    \caption{The adaptation effect averaged across all 75 models when (a) they were adapted to each of the structures and tested on either the same structure (blue, bottom) or different structure (pink, top) and (b)  they were adapted to RCs and tested on non-RCs or vice versa (pink bars); or when they were adapted to RCs or non-RCs and tested on other RCs or and non-RCs respectively (blue bars). Greater values indicate more similarity between adaptation and test structures. Error bars reflect 
    95\% CIs. }
\end{figure*}

When the LM is adapted to sentences such as~\ref{ex1}, it could adjust its expectations about several properties of the sentence, some more linguistically interesting than others. For instance, it could learn that there are three determiners in the sentence, that the third word of the sentence is \textit{that}, that sentences have nine words, that every verb is preceded by a noun, and so on and so forth. If there is a decrease in surprisal when a model is adapted to \ref{ex1} and tested on \ref{ex2}, it is unclear if this is because the model learned to expect object relative clauses or if it learned to expect any of the other mentioned properties. 

To minimize the likelihood that the adaptation effects are driven by irrelevant properties of the sentence,  we introduced several sources of variability to our templates: nouns could either be singular or plural, noun phrases could be optionally modified by an adjective, adjectives were optionally modified with an intensifier and verb phrases were optionally modified with adverbs which could occur either pre-verbally or post-verbally (details in the Supplementary Materials).\footnote{The templates and code for all the analyses along with the data can be found on GitHub: https://github.com/grushaprasad/RNN-Priming}

\subsection{Models}
We used 75 of the LSTM language models trained by \citet{vanschijndel19}; these LMs varied in the number of hidden units per layer (100, 200, 400, 800, 1600) and the number of tokens they were trained on (2 million, 10 million or 20 million). For each training corpus size, \citeauthor{vanSchijndel18adaptation} trained models on five disjoint subsets of the WikiText-103 corpus, to ensure that the results generalized across different training sets. 

\subsection{Calculating the adaptation effect (AE)}
For every structure, we computed the similarity between that structure and every other structure (including itself) as described in  Section~\ref{sec:adapt(y|x)}. This process is schematized in Figure~\ref{fig:exp_schema}. The surprisal values were averaged across the entire sentence.\footnote{Unknown words were excluded from this average.} 

We found that $\mathbb{A}(B\mid A)$ was proportional to the surprisal of $B$ prior to adaptation (see Supplementary Materials). As a consequence, for three structures $X$, $Y$ and $Z$, $\mathbb{A}(Y\mid X)$ could be greater than $\mathbb{A}(Z\mid X)$ merely because $Y$ was a more surprising structure to begin with than $Z$. In order to remove this confound, we first fit a linear regression model predicting $\mathbb{A}(Y\mid X)$ from the surprisal of $Y$ prior to adaptation ($\textit{Surp}(Y)$):

\vspace{-1.8em}
$$\mathbb{A}(Y\mid X) =  \beta_0 + \beta_1 Surp(Y) + \epsilon$$
\vspace{-1.8em}

We then regressed out the linear relationship between  $\mathbb{A}(Y\mid X)$ and $\textit{Surp}(Y)$ as follows:
\vspace{-0.5em}
\begin{align*} 
\textit{AE}{(Y\mid X)} &= \mathbb{A}(Y\mid X) - \beta_1 Surp(Y) \\
& = \beta_0 + \epsilon
\end{align*}
\vspace{-2em}

Since $\textit{Surp}(Y)$ was centered around its mean, $\beta_0$ reflects the mean of $\mathbb{A}(Y\mid X)$ when $\textit{Surp}(Y)$ is equal to the mean surprisal of all sentences prior to adaptation. The term $\epsilon$ reflects any variance in $\mathbb{A}(Y\mid X)$ that is not predicted by $\textit{Surp}(Y)$. By summing these two terms together, $\textit{AE}{(Y\mid X)}$ reflects the change in surprisal for $Y$ after adapting to $X$ that is independent of $\textit{Surp}(Y)$. 

\subsection{Statistical analyses}
We used linear mixed effects models \cite{pinheiro2000mixed} to test for statistical significance; all of the results reported below were highly significant. Details about the statistical analyses can be found in the Supplementary Materials.

\section{Results}

\subsection{Validating AE as a similarity metric}
As discussed in Section~\ref{sec:adaptation_as_priming}, under the adaptation-as-priming paradigm, we would expect sentences that share the same specific structure to be more similar to each other than lexically matched sentences that do not share the structure.\footnote{By lexically matched we mean that all content words were shared between sentences.} In other words, if $X_1$ and $X_2$ are non-lexically-overlapping sets of sentences with shared structure $S_X$, and $Y_2$ is a set of sentences with structure $S_Y$, but is lexically matched with $X_2$, then we would expect $AE(X_2 \mid X_1) > AE(Y_2 \mid X_1)$. We found this prediction to be true for all of our seven structures (Figure~\ref{fig:analysis1}), thus validating our similarity metric. 

\subsection{Similarity between sentences with different types of VP coordination}
Our two coordination conditions were structurally identical to each other but varied in their semantic plausibility --- the sentences in PS/ORC-matched coordination condition were often semantically implausible whereas sentences in ASRC-matched condition were always semantically plausible (see footnote~\ref{footnote:coord_semantics}). If sentences that were structurally similar were close together irrespective of semantic plausibility, then we expect sentences with coordination to be more similar to each other than lexically matched sentences with RCs. Consistent with this prediction, the adaptation effect for models adapted to one type of coordination was greater when the models were tested on sentences with the other type of coordination than when they were tested on sentences with RCs (top panel of Figure~\ref{fig:analysis2}). 

\subsection{\label{section:rc} Similarity between sentences with different types of RCs}
Unlike sentences with coordination, sentences with different types of RCs differ from each other at a surface level (see Table~\ref{tab:ex}). However, at a more abstract level they all share a common property: a gap. If the RNN LMs were keeping track of whether or not a sentence contained a gap, we would expect sentences with different types of RCs to be more similar to each other in the RNN LMs' representation space than lexically matched sentences without a gap. In other words, if $RC_X$ and $RC_Y$ are two different types of RCs and $Coord_Y$ is a sentence with verb coordination lexically matched with $RC_Y$, then we would expect $AE(RC_Y \mid RC_X) > AE(Coord_Y \mid  RC_X)$.

Consistent with this prediction, the adaptation effect for models adapted to RCs was greater when they were tested on sentences with other types of RCs than when they were tested on sentences with coordination (bottom panel of Figure~\ref{fig:analysis2}). This suggests that the LMs do keep track of whether or not a sentence contains a gap, even though this property is not overtly indicated by a lexical item that is shared across all types of RCs. 

\begin{figure}
    \includegraphics[width = 0.4\textwidth]{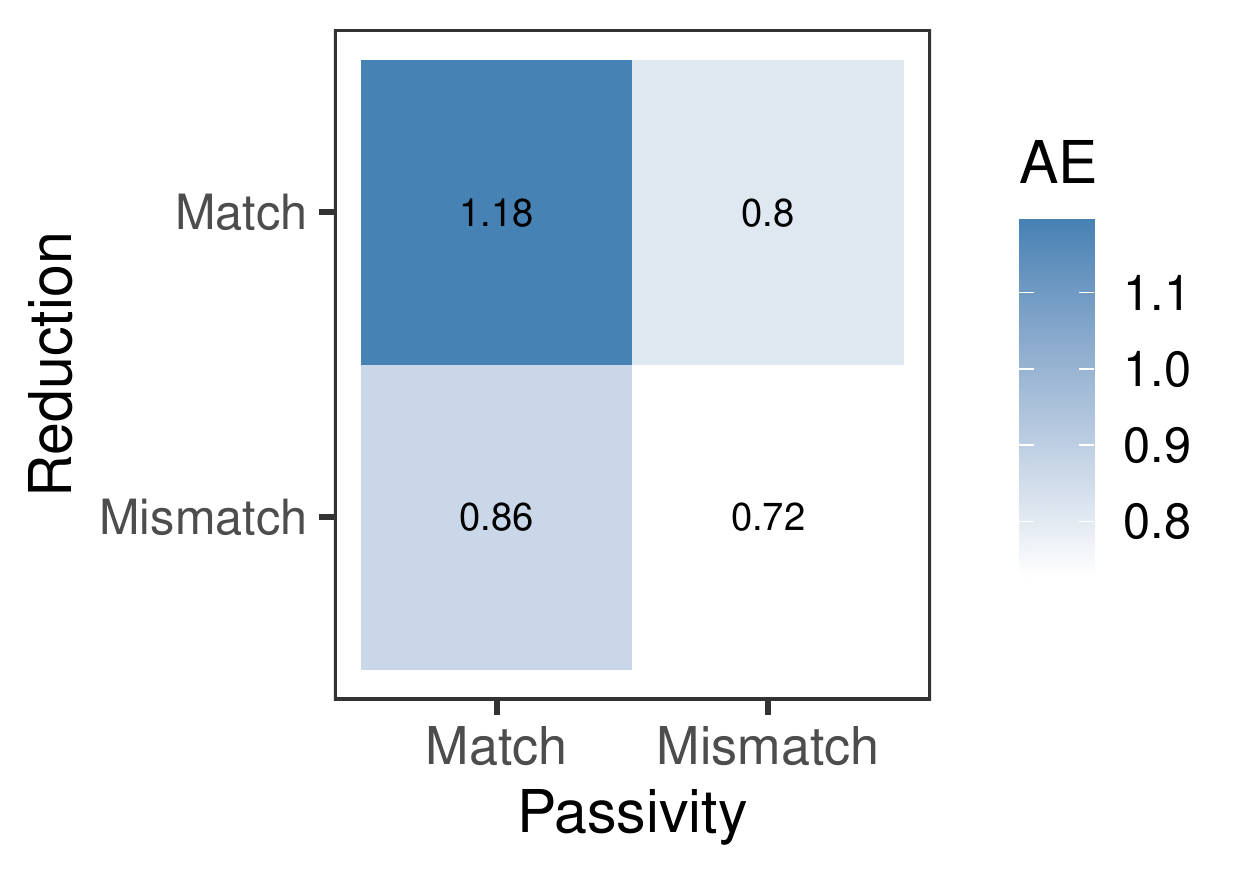}
    \caption{The adaptation effect when models adapted to sentences with reduced and unreduced RCs are tested on sentences that match only in reduction (top right), match only in passivity (bottom right), match in both reduction and passivity (top left) or sentences that match in neither (bottom right).}
    \label{fig:analysis3}
\end{figure}

\subsection{Similarity between sentences belonging to different sub-classes of RCs}
The different types of RCs we tested can be divided into sub-classes based on at least two linguistically interpretable features: reduction and passivity. Reduction distinguishes reduced passive and object RCs on the one hand from unreduced passive and object RCs on the other. Passivity distinguishes reduced and unreduced passive RCs on the one hand from reduced and unreduced object RCs on the other. The LMs could be tracking either, both or none of these features. 

We probed whether the LMs track these features by comparing the similarity between sentences that share one feature but not the other, with the similarity between sentences that share neither feature. If the adaptation effect is greater when there is a match in one feature than when there is a match in neither of the features, we can infer that the LMs track whether sentences have that feature. 
We found that the LMs track both of these features (Figure~\ref{fig:analysis3}).  

Additionally, we probed which of the features contributes more towards the similarity between sentences by comparing the similarity between sentences that match only in passivity with sentences that match only in reduction. When the adaptation and test sets matched only in passivity, the adaptation effect was slightly (but significantly) greater than when the adaptation and test sets matched only in reduction (Figure~\ref{fig:analysis3}). In other words, in the LMs' representation space, \ref{ex:rorc2} is more similar to \ref{ex:uorc2} than it is to \ref{ex:rprc2}, suggesting that passivity contributes more towards the similarity between sentences than reduction.

\ex. \label{ex:rorc2} The conspiracy the employee welcomed divided the country. 

\ex. \label{ex:uorc2} The conspiracy that the employee welcomed divided the country. 

\ex. \label{ex:rprc2} The conspiracy welcomed by the employee divided the country. 

This result is both intuitive and linguistically interpretable --- the edit distance between reduced and unreduced RCs is smaller than the that between object and passive RCs; the syntax tree for \ref{ex:rorc2} is also more similar to \ref{ex:uorc2} than it is to \ref{ex:rprc2}. 

\subsection{\label{section:nhid_csize}What properties of sentences drive the similarity between them?}
Our analyses so far have demonstrated that sentences that belong to linguistically interpretable classes (e.g., sentences that match in reduction) are more similar to each other in the LMs' representation space than they are to sentences that do not belong to those classes (e.g., sentences that do not match in reduction). However, it is unclear what properties of the sentences are driving this similarity between members of the class. For almost all of the linguistically interpretable classes we considered, all sentences belonging to a class shared at least some, if not all, function words. The only exception was the class of all RCs, where the property shared by all sentences in this class (the presence of a gap) was not overtly observable. Therefore, it is possible that the similarity between members of most of the classes we tested was being driven entirely by the presence of these function words.

In order to test whether the similarity between members of classes was indeed being driven by the presence of shared function words, we compared the representation space of the models we tested in the previous sections (henceforth \textit{trained models}) with the representation space of models trained on no data (henceforth \textit{baseline models}). Since the baseline models were only ever exposed to the 20 sentences in the adaptation set and there was no lexical overlap in content words between adaptation and test sets, any similarity between sentences in the representation space of these models would be driven by the presence of function words. If the similarity between sentences in the representation space of the trained models was being driven by factors other than the presence of function words, we would expect this similarity to be greater than the similarity between these sentences in the representation space of the baseline models. 

\begin{figure}
    \includegraphics[width = 0.45\textwidth]{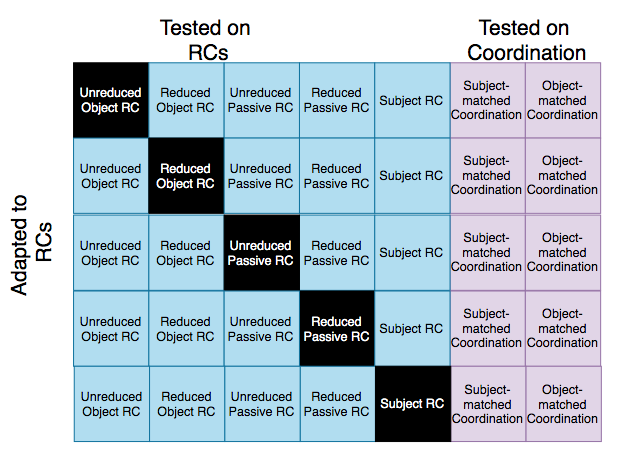}
    \caption{A schematic of how $\mathbb{D}(\textit{RC}, \neg \textit{RC})$ is calculated. For any given row, the black square indicates the specific structure the models were adapted to, the blue squares indicate other structures that belong to the same linguistically defined class as the black square and the pink squares indicate the structures that do not belong to this linguistically defined class. In calculating the distance, we first calculated the proportion between the mean adaptation effect for the blue squares and the mean adaptation effect for pink squares for each row. We then averaged across the proportion for each row to arrive at one number.}
    \label{fig:analysis4_schematic}
\end{figure}

We cannot directly use adaptation effect to compare the similarity between sentences in the representation spaces of trained models and baseline models, however: models trained on more data are likely to have stronger priors and are therefore less likely to drastically change their representations after 20 sentences than models trained on less data. In order to mitigate this issue, we defined a distance measure between sentences that belong to a class and sentences that do not belong to a class $S_X$ as follows (see Figure~\ref{fig:analysis4_schematic} for a schematic):

\vspace{-0.75em}$$\mathbb{D}(S_X, \neg S_X) = \frac{AE(X_2 \mid X_1)}{AE(\neg X_2 \mid X_1)}$$

\vspace{-0.75em}This value would be greater than one if sentences that belonged to a class were more similar to each other than they were to sentences that did not belong to the class. Since the strength of prior belief would affect sentences that belong to the class the same way it would affect sentences that do not belong to the class, the effect would cancel out. 

\begin{figure*}
    \centering
    \begin{subfigure}[b]{0.5\textwidth}
    \centering
        \includegraphics[width=\textwidth]{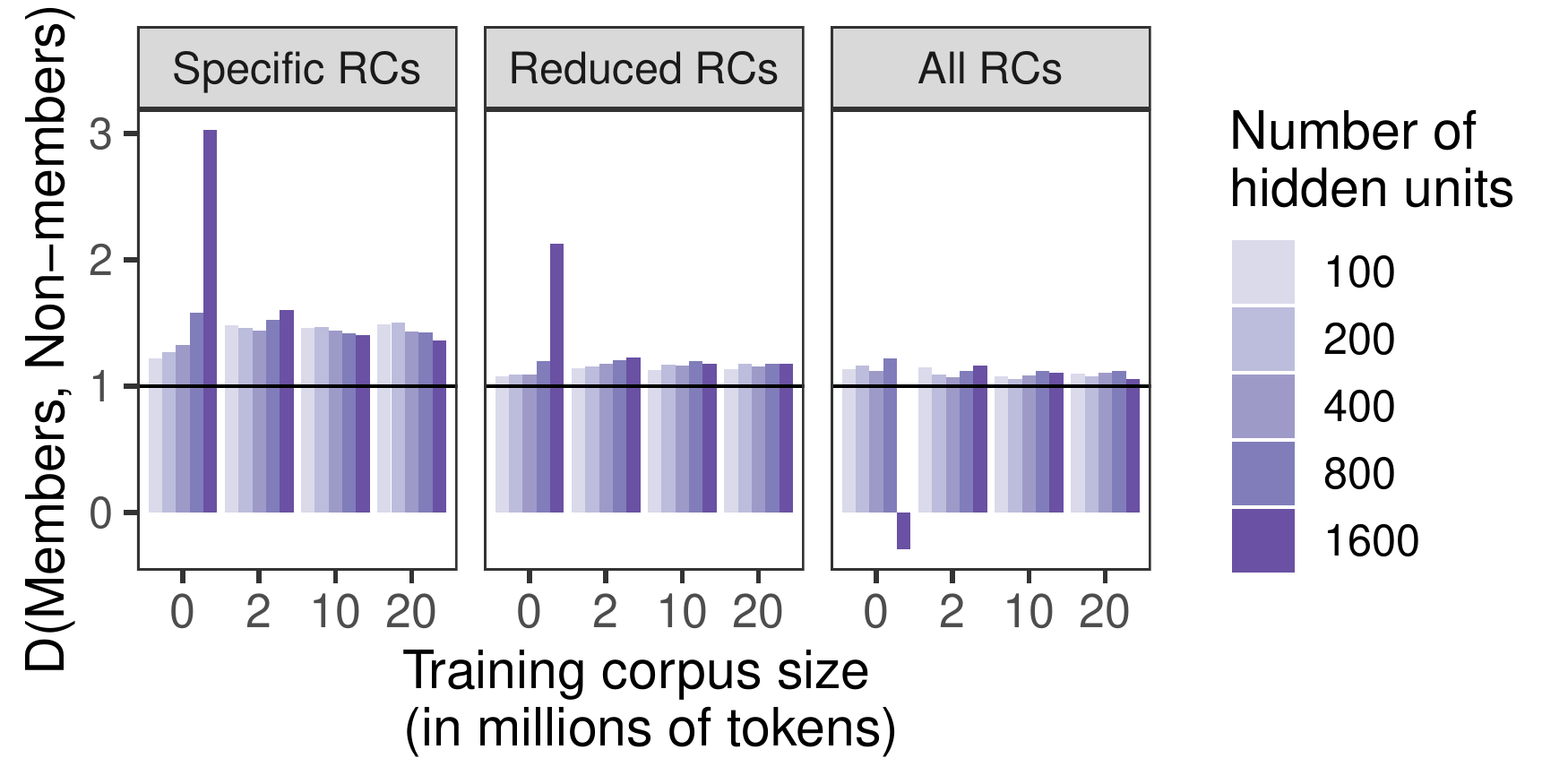}
        \caption{}
        \label{fig:analysis4}
    \end{subfigure}\hfill
    \begin{subfigure}[b]{0.5\textwidth}
        \includegraphics[width=0.85\textwidth]{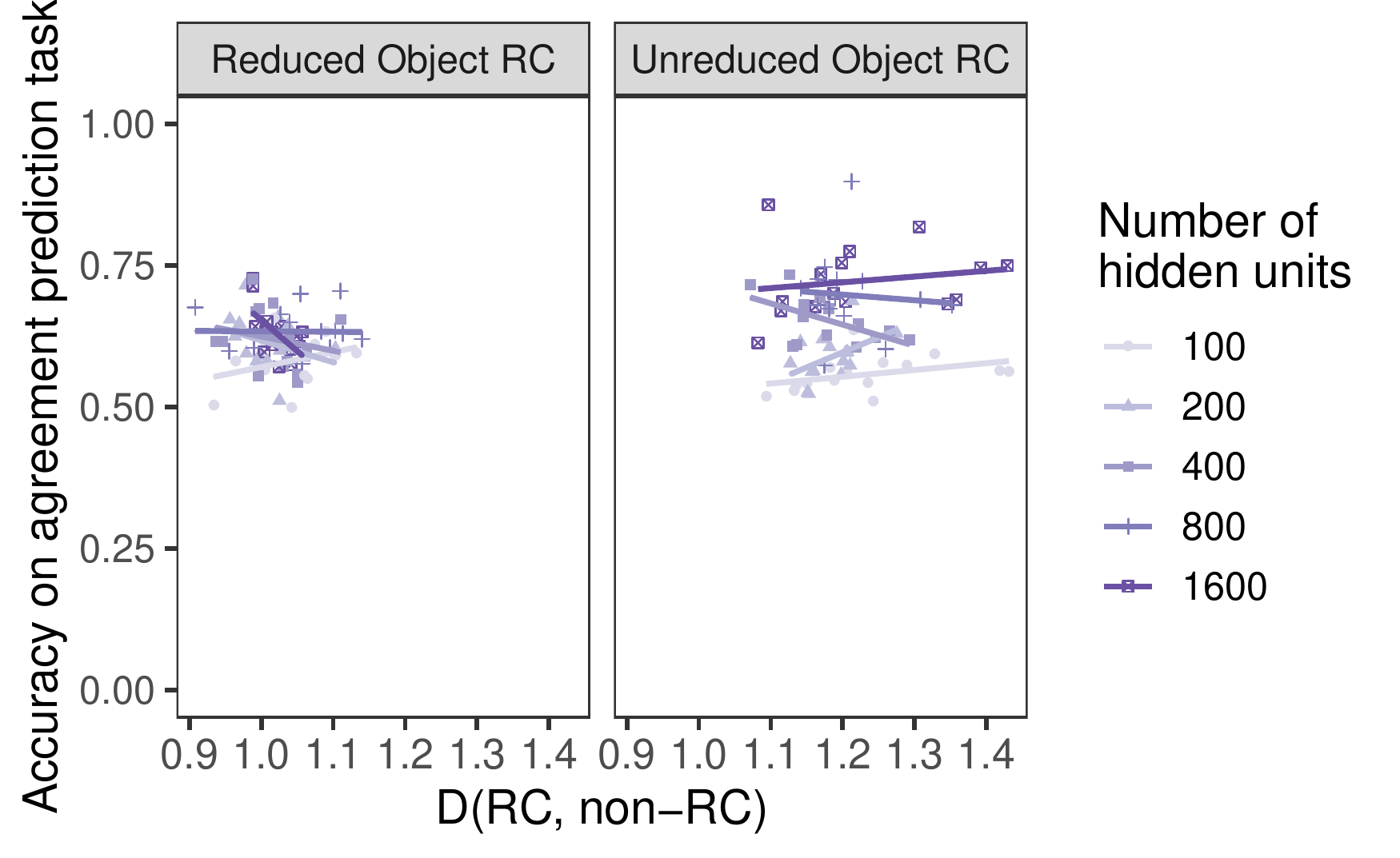}
        \caption{}
        \label{fig:accuracy}
    \end{subfigure}
    \caption{(a) Effect of hidden layer size and corpus size on the distance between sentences with specific RCs and sentences without (left), between sentences that match in reduction and sentences that do not (middle) and between sentences with RCs and sentences without (right). The solid black line indicates the point at which sentences that belong to a particular class are equally similar to other sentences that belong to that class and sentences that do not. (b) Agreement prediction accuracy on reduced object RCs and unreduced object RCs as a function of $\mathbb{D}(RC, \neg RC)$}
\end{figure*}

We measured the distance between members and non-members for three linguistically interpretable classes: sentences which contained the same type of RC, sentences that matched in their reduction or sentences that contained any type of RC. In our baseline models, for all three classes, sentences that belonged to one of these classes were more similar to each other than sentences that did not belong to that class (Figure~\ref{fig:analysis4}). This was surprising for the class of sentences that contained any type of RC because there was no function word that was shared by all sentences in this class. We hypothesize that this is because sentences without RCs always contained the word \textit{and}, whereas sentences with RCs never did.

In cases where members of the class shared at least some function words, the distance between sentences that belonged to the class and sentences that did not for the trained models was greater than that for the baseline models. This suggests that the similarity between sentences in the representation space of trained models was being driven by factors other than the mere presence of function words. However, somewhat surprisingly, as the number of training tokens increased, the distance between members and non-members decreased. 

In the case where the members of the class did not share any function words, the distance between sentences that belonged to the class and sentences that did not belong to the class did not differ between the trained models and the baseline models. This suggests that any similarity between sentences in the representation space of trained models was driven purely by the presence (or in this case absence) of lexical items.

\subsection{Does $\mathbb{D}(\textit{RC}, \neg \textit{RC})$ predict agreement prediction accuracy?}

\citet{marvin18} created a dataset that evaluated the grammaticality of the predictions of language models. Using this dataset, they showed that LSTM LMs could not accurately predict the number of the main verb if the main clause subject was modified by an object RCs (either reduced or unreduced). However, the models had better performance if the main clause was modified by an active subject RC. For example, the models were at near chance levels in predicting that \ref{ex:orcagreement_grammatical} should have higher probability than \ref{ex:orcagreement_ungrammatical}, but were slightly better at predicting that \ref{ex:srcagreement_grammatical} should have higher probability than \ref{ex:srcagreement_ungrammatical}:

\ex. \a. \label{ex:orcagreement_grammatical}\  The farmer that the parents love swims.
\b. \label{ex:orcagreement_ungrammatical}\  *The farmer that the parents love swim. 

\ex.
\a. \label{ex:srcagreement_grammatical}\ The farmer that loves the parents swims. 
\b. \label{ex:srcagreement_ungrammatical} \  *The farmer that loves the parents swim.

\begin{figure*}
    \centering
    \includegraphics[width = 0.65\textwidth]{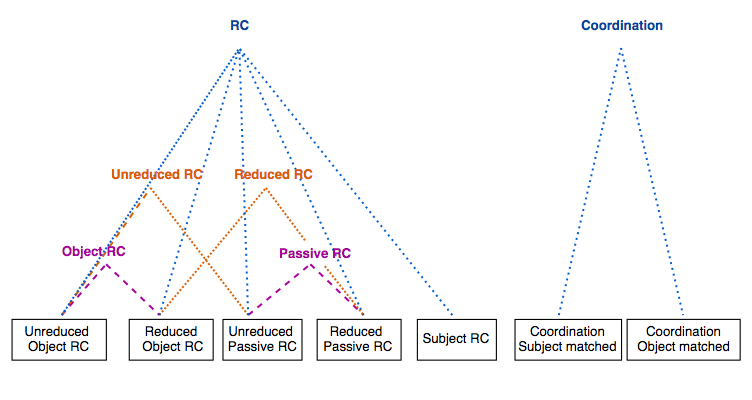}
    \caption{A schematic of how sentences belonging to different linguistically defined classes are related to each other in the LMs' representation space. Each colour indicates a different level of hierarchy.}
    \label{fig:hierarchy}
\end{figure*}

One possible explanation for this poor performance is that object RCs, either reduced or unreduced, are quite infrequent \cite{roland07}. If the LM treats object RCs as unrelated to other RCs, there are likely very few training examples from which the models can learn about subject-verb agreement when the subject is modified by an object RC. If the LM had instead treated object RCs as belonging to the same class as other RCs, it could learn to generalize from training examples of subject-verb agreement when the subject is modified by other RCs. This suggests the hypothesis that agreement prediction accuracy on object RCs will be higher in LMs in which the representation of object RCs is more similar to the representation of other RCs.

The similarity between object RCs and other RCs was defined as in the previous section (the proportion of blue squares to pink squares of the top two rows in Figure~\ref{fig:analysis4_schematic}). There was an increase in accuracy as the number of hidden units increased (see Figure~\ref{fig:accuracy}). However, the similarity between object RCs and other types of RCs did not significantly correlate with agreement prediction; we therefore did not find any evidence for the hypothesis mentioned above.\footnote{Similar patterns were observed for the other constructions in the dataset. See Supplementary Materials.} 

\section{Discussion}

Drawing on the syntactic priming paradigm from psycholinguistics, we proposed a new technique to analyze how the representations of sentences in neural language models (LMs) are organized. Applying this paradigm to sentences with relative clauses (RCs), we found that the representations of these sentences were organized in a linguistically interpretable hierarchical manner (summarized in Figure~\ref{fig:hierarchy}). 

We investigated whether this hierarchical organization was driven by function words that are shared among sentences sentences or whether there was evidence that LMs were tracking more abstract properties of the sentence. We found that for at least some linguistically interpretable classes, sentences that belonged to these classes were more similar to each other in the representation space of the LMs we tested than in the representation space of baseline LMs that were not trained on any data. This suggests that the trained LMs were capable of tracking abstract properties of the sentence. 

However, for linguistically interpretable classes in which sentences shared a non-lexically observable property (e.g. presence of a gap), sentences were as similar to each other in the representation space of the LMs we tested as in the representation space of baseline LMs. Taken together, these results suggest that LMs might be able to track abstract properties of classes of sentences only if these classes also share a lexically observable property. 

Additionally, we found that the sentences belonging to linguistically interpretable classes were more similar to each other in the representation spaces of models trained on 2 million tokens than in the representation spaces for models trained on 20 million tokens. We infer from this that LMs' ability to track abstract properties of sentences decreases with an increase in the training corpus size. This suggests that if we want these LMs to track more abstract linguistic properties, training them on more data from the same distribution is unlikely to help (cf. \citealt{vanschijndel19}). Future work can explore how to bias these models to track linguistically useful properties through architectural biases \cite{dyer16}, training on auxiliary tasks \cite{enguehard17} or data augmentation \cite{perez17}. 

We hypothesized that models' accuracy on subject verb agreement when preceded by object RCs would increase as the similarity between object RCs and the other types of RCs increased. However, we did not find evidence for this. This could either be because the similarity between object RCs and the other types of RCs was too weak to be useful (see Figure~\ref{fig:analysis4}) or because the LMs do not use this property when predicting verb agreement. Future work can disambiguate these reasons by testing models that are biased to treat sentences with object RCs and other RCs as being similar. 

Finally, our method allows us to generate a similarity matrix in the LMs representation space for any given set of structures. In the future, generating a similar matrix for human representations using priming experiments and comparing these two matrices using analysis methods from cognitive neuroscience \cite{kriegeskorte08} may enable us to gain insight into how human-like the LM representations are and vice versa. 

\section{Conclusion}
We proposed a novel technique to analyze how the representations of various syntactic structures are organized in neural language models. As a case study, we applied this technique to gain insight into the representations of sentences with relative clauses in RNN language models and found that the representations of sentences were organized in a linguistically interpretable manner. 

\section{Acknowledgments}
We would like to thank Sadhwi Srinivas and the members of the CAP lab at JHU for helpful discussions and valuable feedback.

\bibliography{RNNPriming_CONLL2019_arxiv}
\bibliographystyle{acl_natbib}

\newpage

\appendix
\section{Templates}
	We created seven templates (one for each of the structures we tested) to generate the adaptation and test sets. Each template had seven slots: subject, object of the relative clause, object of the main clause, verb in the relative clause, verb in the main clause, adverb for the main clause and adverb for the relative clause. The adverb arguments were blank strings half the time. The seven templates varied in the order in which they combined these arguments together to form a sentence. Therefore, for a given set of arguments, we were able to generate seven lexically matched sentences with different structures. 
	
	We included several sources of noise in our sentence generation process. 
	\begin{itemize}
	    \item Each noun slot was filled by a plural noun 40\% of the time. 
	    \item Every noun phrase was modified with an adjective with 50\% probability and every adjective was further modified with an intensifier with 40\% probability. 
	    \item In cases when a verb (in the main clause or relative clause) was modified by an adverb, the adverb occurred pre-verbally or post-verbally with equal probability. 
	\end{itemize}
	
	The slots in the templates were filled by 223 verbs, 164 nouns, 24 adverbs and 78 adjectives. In order to ensure semantic plausibility, we created sub-classes of nouns, adverbs and adjectives and manually specified which sub-classes could combine together. For example, the noun subclass ``human" consisted of the nouns \textit{friend}, \textit{cousin}, \textit{partner}, \textit{sibling} and \textit{colleague}. This class could serve as subjects for 38 verbs and could be modified by four sub-classes of adjectives. Similarly the verb \textit{congratulated} could take the noun subclass ``human" as its subject and the noun subclasses ``scienceperson" and ``power" and as its object (e.g., \textit{scientist}, \textit{researcher} etc.; \textit{principal}, \textit{manager} etc.). Additionally, it could be modified by adverb subclasses ``sad" and ``time" (e.g, \textit{sadly},\textit{gloomily} etc.; \textit{yesterday}, \textit{last month} etc.) 
	
	We ensured that there was no lexical overlap between adaptation and test sets, apart from function words (like \textit{the}, \textit{and}, \textit{by}, \textit{that} etc) and intensifiers (like \textit{very}, \textit{rather}, \textit{quite etc}). We also ensured that verbs, nouns, adverbs and adjectives were not repeated within the same sentence. 
	
	\newpage
	\section{Relationship between $\mathbb{A}(Y\mid X)$ and $\textit{Surp}(Y)$ prior to adaptation}
	\begin{figure}[h]
        \centering
        \includegraphics[width = 0.4\textwidth]{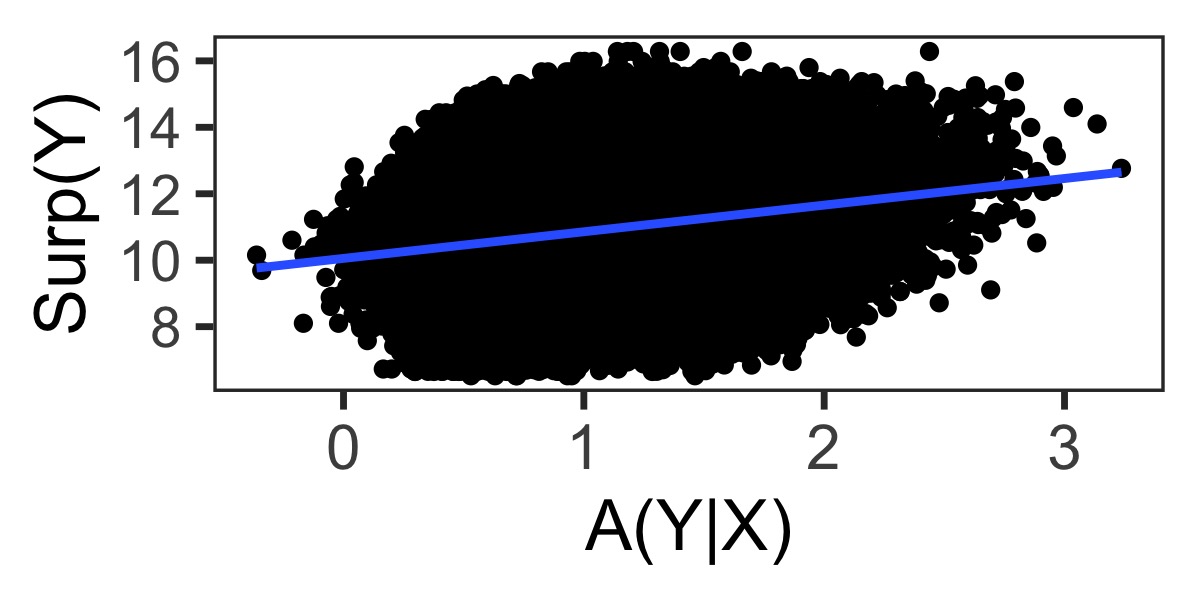}
        \caption{}
    \end{figure}
    
    \noindent\textbf{LM formula:} \newline
    $A(Y \mid X) \sim center(\textit{Surp}(Y))$ \newline
    
    \noindent\textbf{Results}$: \newline
    \hat{\beta} = 0.061, \textit{SE} = 0.0003, p < 2e-16$
    
    \section{Statistical Analyses:}
	This section contains details about the statistical analyses for all the results described in the main paper. In describing the formula for our mixed effects models we use standard LMER notation. 
	
	\subsection{Validating AE as a similarity metric}
	
	For this analyses we fit a separate LMEM for each of the different structures that models could get adapted to. \newline
	
	\noindent \textbf{LMER formula:} \newline
	AE $\sim$ structure + (1 $\mid$ adaptlist) + (1 $\mid$ clist) \vspace{-0.5em}
	
	\begin{itemize}
	    \item Structure is a categorical variable coded as $1$ if the test structure is the same as the adaptation structure and $-1$ if it is different. \vspace{-0.5em}
	    \item adaptlist: Which of the 10 adaptation-test sets we generated was the model adapted to and tested on? 
	    \item clist: Which subset of Wikipedia was the model trained on?
	\end{itemize}\vspace{-0.5em}
	
\begin{table}[h!]
    \centering
    \resizebox{0.48\textwidth}{!}{
    \begin{tabular}{llll}
        \toprule
        Structure adapted to & $\hat{\beta_{structure}}$ &  \textit{SE}  & p-value \\
        \midrule
        Unreduced Object RC & 0.256 & 0.001 & $p<2e-16$ \\
        Reduced Object RC & 0.171 & 0.001 & $p<2e-16$ \\
        Unreduced Passive RC & 0.229 & 0.001 & $p<2e-16$ \\
        Reduced Passive RC & 0.100 & 0.001 & $p<2e-16$ \\
        Active Subject RC & 0.194 & 0.001 & $p<2e-16$ \\
        Subject coordination & 0.147 & 0.001 & $p<2e-16$ \\
        Object coordination & 0.145 & 0.001 & $p<2e-16$ \\
        \bottomrule
    \end{tabular}
    }
\end{table}

\subsection{Similarity between sentences with different types of VP coordination}
	We fit the following mixed effect model on LMs that were adapted to sentences with coordination.  
	
	\noindent \textbf{LMER formula:} \newline
	AE $\sim$ testtype + (1 $\mid$ adaptlist) + (1 $\mid$ clist)
	
	testtype was a categorical variable coded as $1$ if the model was tested on sentences with RCs and $-1$ if the model was tested on sentences with the other type of coordination (e.g, for model adapted to ASRC-matched coordination, testtype was $-1$ if it was tested on PS/ORC-matched coordination)
	
	$\hat{\beta} = -0.173, \textit{SE} = 0.0007, p < 2e-16$
	
	\subsection{Similarity between sentences with different types of RCs}
	We fit the following mixed effect model on LMs that were adapted to sentences with RCs.  
	
	\noindent \textbf{LMER formula:} \newline
	AE $\sim$ testtype + (1 $\mid$ adaptlist) + (1 $\mid$ clist)
	
	testtype was a categorical variable coded as $1$ if the model was tested on sentences with other types RCs (e.g., for a model adapted to unreduced object RC, the value of testttype was $1$ when tested on reduced object RC, reduced/unreduced passive RC and active subject RC). It was coded as $-1$ if the model was tested on sentences with coordination. 
	
	$\hat{\beta} = 0.038, \textit{SE} = 0.0004, p < 2e-16$
	
	\subsection{Similarity between sentences belonging to different sub-classes of RCs}
	
	We fit the following mixed effect model on LMs that were adapted to sentences with object or passive RCs.  
	
	\noindent \textbf{LMER formula:} \newline
	AE $\sim$ testtype + (1 $\mid$ adaptlist) + (1 $\mid$ clist)
	
	testtype was a categorical variable with four levels: passive match, reduced match, no match and both match. Since there were four levels, there were three contrasts. Passive match was chosen as the baseline and coded as $0$ for all of the contrasts. For each contrast, one of the other levels was coded as $1$ --- i.e. in each contrast, the mean adaptation effect of passive match was compared to the mean adaptation effect of one of the other conditions. 
	
	\begin{table}[h!]
    \centering
    \resizebox{0.48\textwidth}{!}{
    \begin{tabular}{llll}
        \toprule
        Contrast & $\hat{\beta_{testtype}}$ &  \textit{SE}  & p-value \\
        \midrule
        Reduced match  & -0.058 & 0.001 & $p<2e-16$ \\
        Both match  & 0.171 & 0.001 & $p<2e-16$ \\
        No match  & -0.143 & 0.001 & $p<2e-16$ \\
        \bottomrule
    \end{tabular}
    }
    \caption{Analysis 5.4}
\end{table}

\subsection{What properties of sentences drive the similarity between them?}
	
	We a separate mixed effects model for each of the three linguistically interpretable classes discussed in Section~5.5 of the paper. We did not include the baseline models in these analyses. \newline
	
	\noindent\textbf{LMER formula}: \newline
	$\mathbb{D}(S, \neg S) $ $\sim$ scale(nhid) * scale(csize) + (1 $\mid$ adaptlist) + (1 $\mid$ clist) 
	
	nhid refers to the number of hidden units (100, 200, 400, 800, 1600) and csize refers to the training corpus size in millions of tokens (2, 10, 20).

	\begin{table}[h]
    \centering
    \resizebox{0.45\textwidth}{!}{
    \begin{tabular}{llll}
        \toprule
        Predictor & $\hat{\beta_{testtype}}$ &  \textit{SE}  & p-value \\
        \midrule
        nhid & 0.008 & 0.002 & $p= 0.003$ \\
        csize & -0.011 & 0.001 & $p= 0.00002$ \\
        nhid:csize & -0.012 & 0.001 & $p= 0.00001$ \\
        \bottomrule
    \end{tabular}
    }
    \caption{$\mathbb{D}(RC, \neg RC)$}
\end{table}

\begin{table}[h]
    \centering
    \resizebox{0.45\textwidth}{!}{
    \begin{tabular}{llll}
        \toprule
        Predictor & $\hat{\beta_{testtype}}$ &  \textit{SE}  & p-value \\
        \midrule
        nhid & 0.016 & 0.001 & $p<2e-16$ \\
        csize & -0.006 & 0.001 & $p= 0.00004$ \\
        nhid:csize & -0.008 & 0.001 & $p < 0.00001$ \\
        \bottomrule
    \end{tabular}
    }
   \caption{$\mathbb{D}(\textit{Reduced match}, \neg \textit{Reduced match})$}
   
    \label{tab:analysis5.52}
\end{table}
	
	\begin{table}[h!]
	\resizebox{0.45\textwidth}{!}{
    \centering
    \begin{tabular}{llll}
        \toprule
        Predictor & $\hat{\beta_{testtype}}$ &  \textit{SE}  & p-value \\
        \midrule
        nhid & -0.007 & 0.002 & $p = 0.008$ \\
        csize & -0.023 & 0.002 & $p<2e-16$ \\
        nhid:csize & -0.040 & 0.001 & $p<2e-16$ \\
        \bottomrule
    \end{tabular}
    }
    
     \caption{$\mathbb{D}(\textit{RC}_X, \textit{RC} \neq X)$}
\end{table}

\subsection{Does $\mathbb{D}(RC, \neg RC)$ predict agreement prediction accuracy?}

We fit a separate linear regression model for LMs adapted to either reduced or unreduced Object RCs. 

\noindent\textbf{LM formula:} \newline
accuracy $\sim$ $\mathbb{D}(\textit{RC}, \neg\textit{RC})$ + scale(nhid) + scale(csize)

\begin{table}[h!]
    \centering
    \resizebox{0.45\textwidth}{!}{
    \begin{tabular}{llll}
        \toprule
        Predictor & $\hat{\beta_{testtype}}$ &  \textit{SE}  & p-value \\
        \midrule
        $\mathbb{D}(\textit{RC}, \neg \textit{RC})$& -0.007 & 0.098 & $p = 0.947$ \\
        nhid & 0.057 & 0.007 & $p\ll 0.0000001$ \\
        csize & 0.001 & 0.008 & $p = 0.879$ \\
        \bottomrule
    \end{tabular}
    }
     \caption{Models adapted to unreduced object RCs}
\end{table}

\begin{table}[h!]
    \centering
    \resizebox{0.45\textwidth}{!}{
    \begin{tabular}{llll}
        \toprule
        Predictor & $\hat{\beta_{testtype}}$ &  \textit{SE}  & p-value \\
        \midrule
        $\mathbb{D}(\textit{RC}, \neg \textit{RC})$& -0.084 & 0.113 & $p = 0.465$ \\
        nhid & 0.013 & 0.005 & $p =  0.018$ \\
        csize & -0.004 & 0.008 & $p = 0.489$ \\
        \bottomrule
    \end{tabular}
    }
     \caption{Models adapted to reduced object RCs}
\end{table}

\section{Relationship between $\mathbb{D}(RC, \neg RC)$ and agreement prediction accuracy for other structures}
 
 \begin{figure}[h!]
        \centering
        \includegraphics[width = 0.47\textwidth]{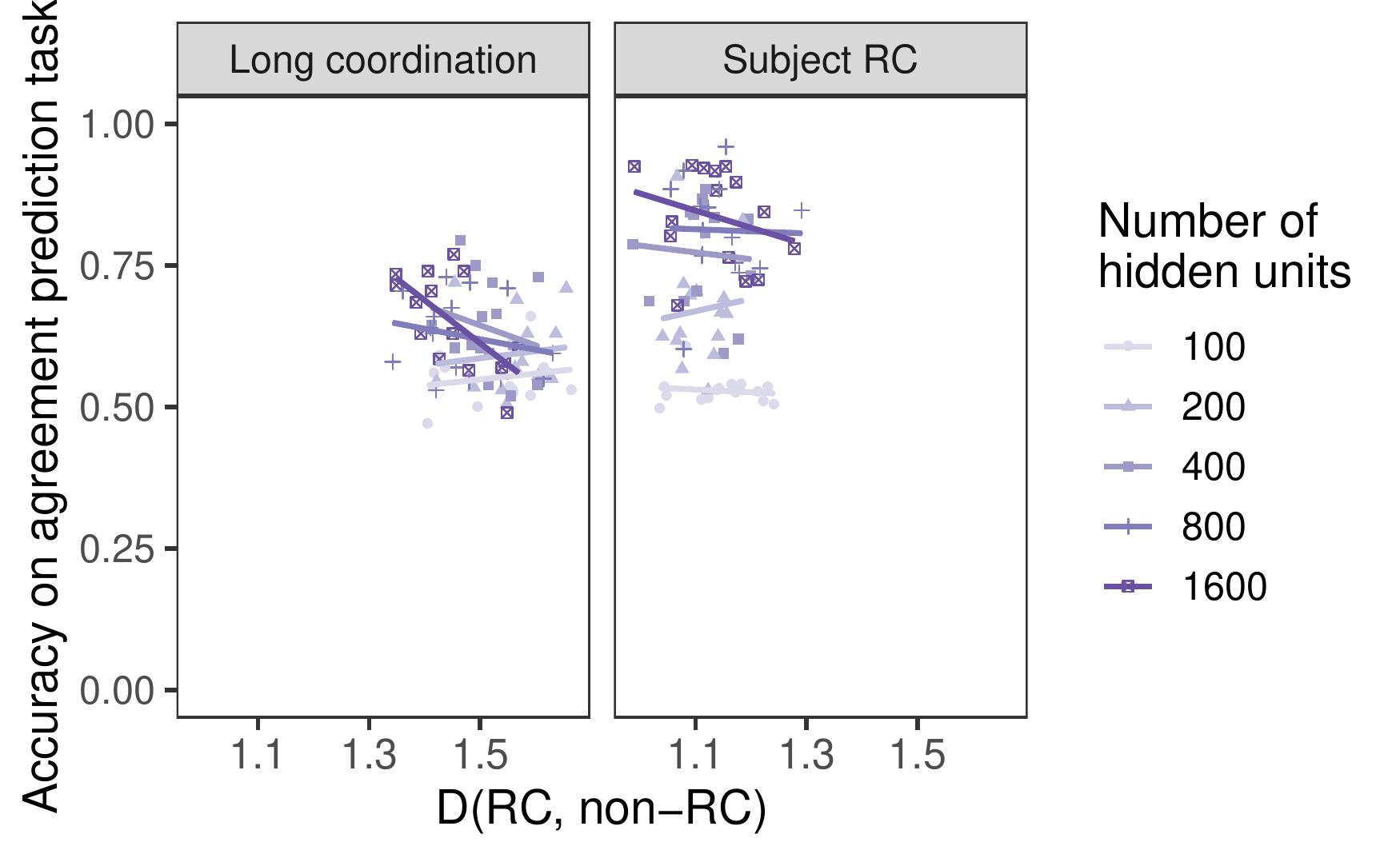}
        \caption{}
    \end{figure}

\noindent\textbf{LM formula:} \newline
accuracy $\sim$ $\mathbb{D}(\textit{RC}, \neg\textit{RC})$ + scale(nhid) + scale(csize)

\begin{table}[h!]
    \centering
    \resizebox{0.45\textwidth}{!}{
    \begin{tabular}{llll}
        \toprule
        Predictor & $\hat{\beta_{testtype}}$ &  \textit{SE}  & p-value \\
        \midrule
        $\mathbb{D}(\textit{RC}, \neg \textit{RC})$& -0.215 & 0.204& $p = 0.297$ \\
        nhid & 0.089 & 0.013 & $p\ll 0.0000001$ \\
        csize & 0.016 & 0.013 & $p = 0.211$ \\
        \bottomrule
    \end{tabular}
    }
     \caption{Models adapted to unreduced active subject RCs}
\end{table}

\begin{table}[h!]
    \centering
    \resizebox{0.45\textwidth}{!}{
    \begin{tabular}{llll}
        \toprule
        Predictor & $\hat{\beta_{testtype}}$ &  \textit{SE}  & p-value \\
        \midrule
        $\mathbb{D}(\textit{RC}, \neg \textit{RC})$& -0.125 & 0.110& $p = 0.259$ \\
        nhid & 0.023 & 0.008 & $p = 0.014$ \\
        csize & 0.025 & 0.008 & $p = 0.003$ \\
        \bottomrule
    \end{tabular}
    }
     \caption{Models adapted to unreduced sentences with long coordinaiton}
\end{table}

\end{document}